\documentclass[letterpaper, 10 pt, conference]{ieeeconf}  %

\IEEEoverridecommandlockouts                              %

\usepackage{pifont}
\newcommand{\cmark}{\ding{51}}%
\newcommand{\xmark}{\ding{55}}%

\usepackage[table,xcdraw]{xcolor} %
\newcommand{\gr}{\rowcolor[gray]{.9}} %
\usepackage{tabularx}
\usepackage{bm}
\usepackage{tikz}
\usepackage{booktabs}
\usepackage{amssymb}
\usepackage{amsmath}
\usepackage{multirow}
\usepackage{xspace}
\usepackage[nospace]{cite}
\usepackage{hyperref}
\usepackage{gradient-text}

\usepackage[capitalize]{cleveref}
\crefname{section}{Sec.}{Secs.}
\Crefname{section}{Section}{Sections}
\Crefname{table}{Table}{Tables}
\crefname{table}{Tab.}{Tabs.}

\definecolor{color_h}{RGB}{34	202	162	}
\definecolor{color_y}{RGB}{30,166,174}
\definecolor{color_D}{RGB}{27, 148, 196}
\definecolor{color_R}{RGB}{23,127,215}
\definecolor{color_a}{RGB}{18,101,222}

\newcommand{\hcolor}[1]{\textbf{\gradientRGB{#1}{30, 231, 161}{15, 120, 248}}}

\title{\LARGE \bf
Unleashing \hcolor{HyDRa}: \textcolor{color_h}{H}\textcolor{color_y}{y}brid Fusion, \textcolor{color_D}{D}epth Consistency and \textcolor{color_R}{R}\textcolor{color_a}{a}dar\\ for Unified 3D Perception}

\author{Philipp Wolters$^{1,2}$\qquad
Johannes Gilg$^{1}$\qquad
Torben Teepe$^{1}$\qquad
Fabian Herzog$^{1}$\qquad\\ %
Anouar Laouichi$^{1}$\qquad
Martin Hofmann$^{2}$\qquad
Gerhard Rigoll$^{1}$\qquad%
\thanks{$^{1}$ Technical University of Munich, Germany}%
\thanks{$^{2}$ FusionRide Technology, Munich, Germany}
}

\begin{document}

\maketitle
\thispagestyle{empty}
\pagestyle{empty}

\newcommand{\sota}{state-of-the-art}

\newcommand{\name}{HyDRa}
\newcommand{\hydra}{HyDRa}
\newcommand{\hatl}{Height Association Transformer}
\newcommand{\hats}{HAT}
\newcommand{\rdc}{Radar-weighted Depth Consistency}
\newcommand{\rdcs}{RDC}

\makeatletter
\DeclareRobustCommand\onedot{\futurelet\@let@token\@onedot}
\def\@onedot{\ifx\@let@token.\else.\null\fi\xspace}
\def\cf{\emph{cf}\onedot} \def\Cf{\emph{Cf}\onedot}
\makeatother

\begin{abstract}
    Low-cost, vision-centric 3D perception systems for autonomous driving have made significant progress in recent years, narrowing the gap to expensive LiDAR-based methods. 
    The primary challenge in becoming a fully reliable alternative lies in robust depth prediction capabilities, as camera-based systems struggle with long detection ranges and adverse lighting and weather conditions.
    In this work, we introduce \name, a novel camera-radar fusion architecture for diverse 3D perception tasks. %
    Building upon the principles of dense Bird's-Eye-View (BEV)-based architectures, \name\ introduces a hybrid fusion approach to combine the strengths of complementary camera and radar features in two distinct representation spaces.
    Our \hatl\ module leverages radar features already in the perspective view to produce more robust and accurate depth predictions.
    In the BEV, we refine the initial sparse representation by a \rdc.
    \name\ achieves a new \sota\ for camera-radar fusion of 64.2~NDS (+1.8) and 58.4 AMOTA (+1.5) on the public nuScenes dataset.
    Moreover, our new semantically rich and spatially accurate BEV features can be directly converted into a powerful occupancy representation, beating all previous camera-based methods on the Occ3D benchmark by an impressive 3.7 mIoU.
    Code and models are available at \href{https://github.com/phi-wol/hydra}{https://github.com/phi-wol/hydra} .

\end{abstract}

\section{Introduction}
\label{sec:intro}
Reliable and affordable 3D perception is an important cornerstone for safe and efficient operation of autonomous vehicles, enabling agents to navigate dynamic environments and complex scenarios.
In this rapidly evolving field, deep learning-based Camera-LiDAR fusion has emerged as the de-facto standard for 3D reconstruction from multiple sensors~\cite{yang2022deepinteraction, liu2022bevfusion, man2023bev, liang2022bevfusion, ding20233dmotformer}. %

While delivering highly accurate geometry sensing, LiDAR sensors still pose a high-cost barrier for large-scale deployment. 
In order to democratize access to autonomous driving systems, LiDAR-free, and vision-centric systems have gained significant traction in recent years and months \cite{li2022bevdepth, Wang2023streampetr, tong2023scene}.  
However, the main practical challenge lies in preserving the spatial information after 2D projection, leading to unstable depth estimation and localization errors \cite{li2022bevdepth,park2022time}.
Another difficulty of passive sensors is posed by dynamic objects that are occluded, yet remain relevant for safe decision making. 

Still underrepresented in this research domain is the incorporation of radar sensors.
Especially camera-radar fusion has been neglected by the academic community \cite{harley2022simple}.
As low-frequency active sensors based on time-of-flight principles, they exhibit a remarkable resilience to adverse weather and lighting conditions, offering metric measurements for perception ranges up to 300 meters \cite{yao2023radarsurveycomprehensive}.
Furthermore, by leveraging the Doppler effect, radar reflections can yield valuable velocity information about moving and occluded objects. %
Combining the two sensor types is key to unlocking the full potential of accessible 3D perception systems \cite{srivastav2023radarsurvey}.

\begin{figure}[t]
	\centering
	\includegraphics[width=0.8\linewidth]{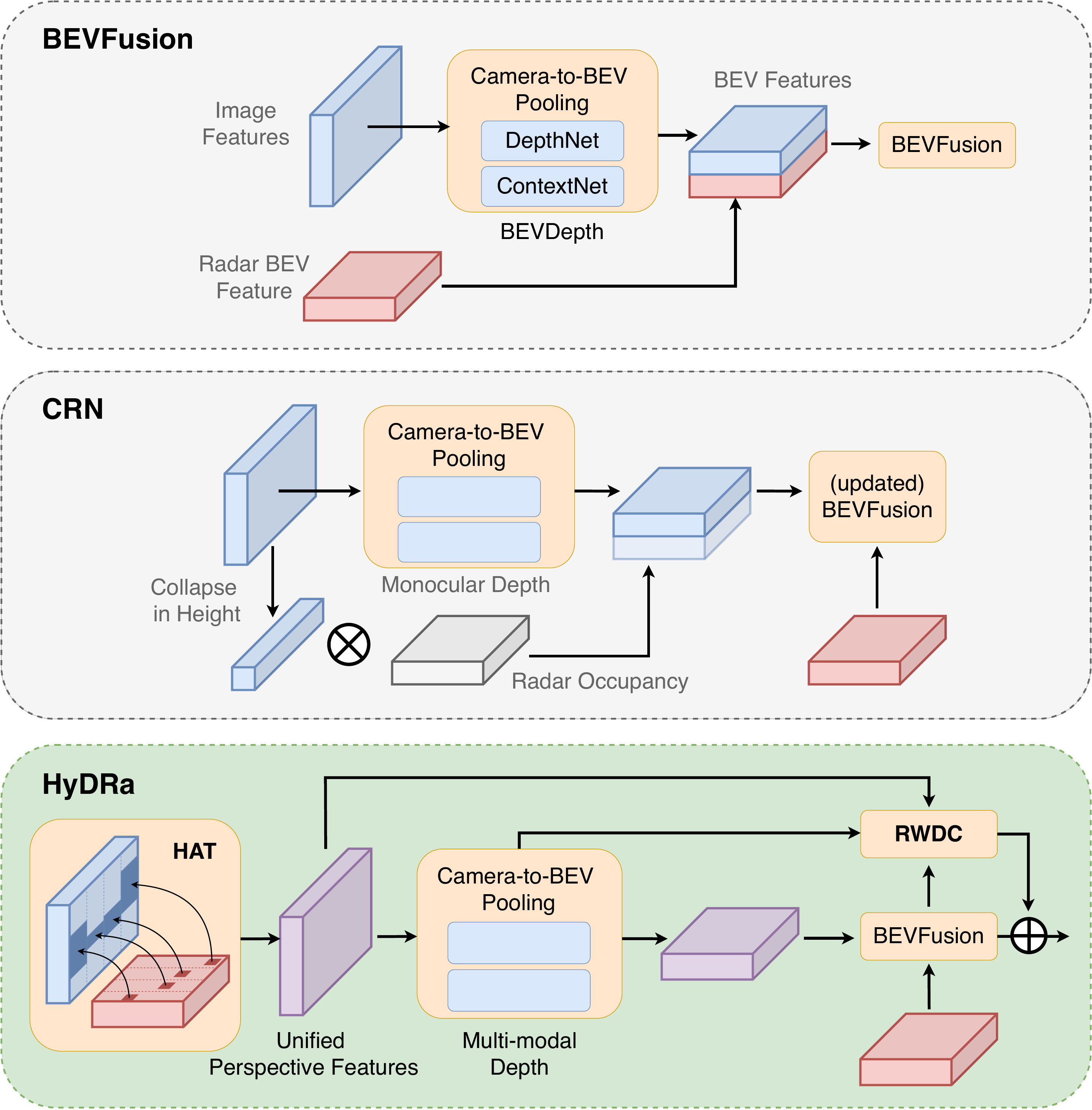}

	\caption{Bridging the view disparity in BEVFusion~\cite{liang2022bevfusion}, CRN~\cite{kim2023crn}, and our HyDRa. We leverage multi-modal feature fusion alraedy for depth splatting.
	}
	\label{fig:comp}
\end{figure}

\begin{figure*}[t]
    \centering
    \begin{tikzpicture}
        \node[anchor=south west,inner sep=0] (mainImage) at (0,0) {\includegraphics[width=0.8\textwidth]{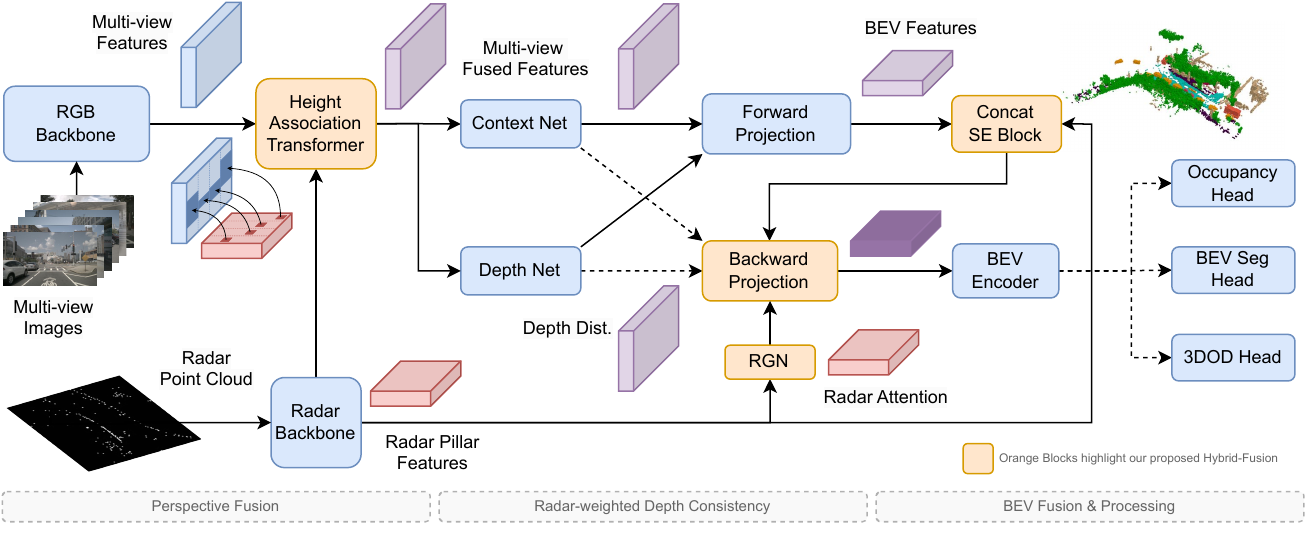}};
        \begin{scope}[x={(mainImage.south east)},y={(mainImage.north west)}]
			\fill [white] (0.84,0.73) rectangle (1,1); 
			\fill [white] (0.813,0.8) rectangle (1,1);
            \node[anchor=north east,inner sep=0] at (0.98*1.05,0.98*1.03) {\includegraphics[scale=0.2]{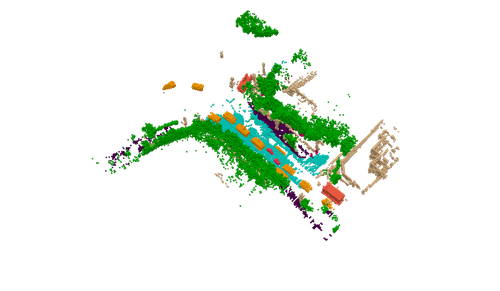}};
        \end{scope}
    \end{tikzpicture}
    
	\caption{\textbf{Architecture of \hydra:} 
	The modality-specific features are fused in two representation spaces: Perspective View and BEV-Space.~1.~The radar features are associated with the image features by the Height Association Transformer.
	With the resulting radar-informed dense depth, the forward projection module generates a sparse BEV representation.
	2. The splatted semantic BEV features and radar-BEV features are concatenated and fused.~ 
	3.~A~depth-aware backward projection refines this representation, guided by radar attention weights before being distributed to task-specific heads.
	}
	\label{fig:teaser}
\end{figure*}

Nevertheless, effectively associating radar detections with camera features is complicated by the unique characteristics of the radar data.
The absence of elevation information, the sparse nature of the radar returns, the presence of noise from multi-path reflections, and the overall high measurement uncertainty make radar data difficult to process and associate with camera features \cite{kim2023crn, srivastav2023radarsurvey}. %
Consequently, the direct application of \sota~LiDAR-like architectures has proven to be ineffective \cite{nabati2021centerfusion}.
CMT \cite{yan2023cmt}, among the current leading methods on the nuScenes benchmark, degrades in performance when incorporating radar points into their model. 
For this reason, we see the need for a significant shift in fusion paradigms when adopting radar. 

Modern practices have focused on adaptive fusion in the BEV, reducing spatial misalignment during feature aggregation \cite{liu2022bevfusion,kim2023crn}.
Despite these improvements, they still struggle with robust depth estimation and rely partly on monocular depth cues for feature transformation \cite{li2022bevdepth} (\cf \cref{fig:comp}).
Existing techniques do not fully capitalize on radar's potential for unified robust depth sensing, often resulting in a dual-stage projection process where each modality stage relies on its own, potentially inconsistent, depth estimation.
We argue that to fully unlock the potential, we must move the fusion stage even earlier.

In this work, we propose \textbf{\hydra}, a state-of-the-art camera-radar fusion architecture well equipped to tackle a variety of 3D perception tasks.
Our contributions are the following: %

\begin{enumerate}
    \item   We introduce our \textbf{\hatl\ (HAT)} module and address the limitations of previous BEV-generating depth networks. 
            By creating unified geometry-aware features in the perspective view, we significantly reduce the translation error.
    \item   We establish a \textbf{\rdc}\ \textbf{(RDC)} for enhancing sparse features in the BEV, tackling misaligned features and occluded objects.
    \item   We extensively ablate our design choices on the \textbf{nuScenes} and \textbf{View-of-Delft} detection and tracking challenges and show that the synergy of all components is crucial for achieving best-in-class results.
    \item   \name\ establishes a pioneering model for camera-radar based 3D \textbf{semantic occupancy prediction} on the Occ3D benchmark \cite{tian2024occ3d}, pushing the upper bound of \sota\ low-cost vision-centric methods. 
\end{enumerate}

\section{Related Work }
\label{sec:related_work}

\subsection{Camera-based Architectures} %

Shifting from perspective-view 3D object detection \cite{mousavian20173d, simonelli2019disentangling, wang2021fcos3d} to Birds-Eye-View (BEV) based perception \cite{philion2020lift, reading2021categorical}, the BEV-based architectures have become the new standard for 3D object detection and tracking \cite{li2023delving, teepe2024earlybird}.
With the introduction of differentiable lifting methods, the image-based output of a standard 2D feature extractor \cite{he2016deep,liu2022convnet, dosovitskiy2021vit,liu2021swin} is transformed into the more downstream task-friendly BEV ground plane representation \cite{zhou2019objects, teepe2024lifting, hu2023planning}.
The BEVDet series \cite{huang2021bevdet, huang2022bevdet4d, huang2022bevpoolv2} and following works \cite{liu2022bevfusion,li2022bevdepth, park2022time} have raised the bar for an efficient and powerful forward projection pipeline, predicting an explicit, dense, and pixel-wise depth distribution and pooling the camera-frustum features in the BEV space, while BEVFormer \cite{li2022bevformer} introduces a spatial cross attention \cite{vaswani2017attention, zhu2021deformable} to implicitly model the 3D relations of regions of interest.
FB-BEV \cite{li2023fb} combines both approaches, addressing the sparsity and inaccuracy inherent in both projection types by ensuring a Depth Consistency between the two modules. 

We will build upon this concept and extend it to the multi-modal domain by introducing a novel \rdc\ and incorporating radar in complementary representation spaces to enforce a more accurate depth estimation.

\subsection{Multi-modal Architectures}
\begin{figure*}[t]
    \centering
    \begin{minipage}[b]{0.48\textwidth}
        \vspace{7.5mm} %
        \includegraphics[width=\linewidth]{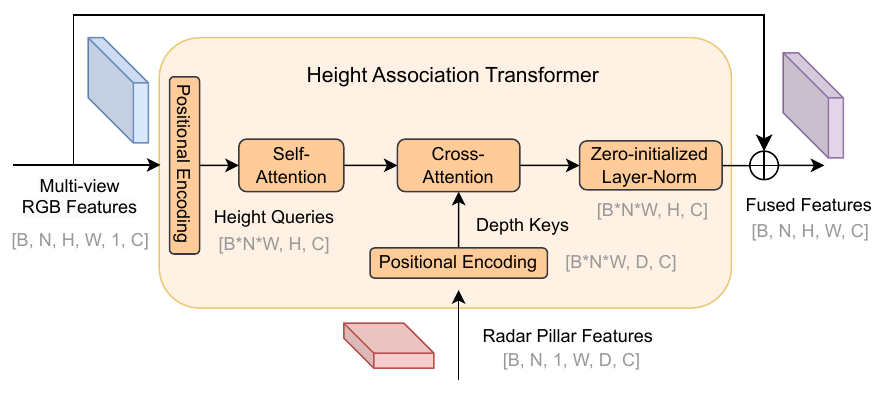}
        \caption{Overview of the \hatl. The radar fusion module exertes a pushing effect into the BEV.}
        \label{fig:hat_details}
    \end{minipage}
    \hfill
    \begin{minipage}[b]{0.48\textwidth}
        \includegraphics[width=0.9\linewidth]{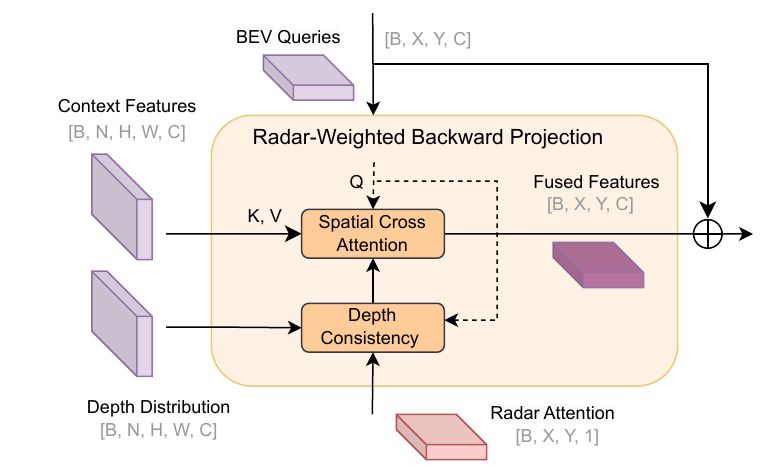}
        \caption{Details of the Radar-weighted Backward Projection. The radar features pull RGB information for refinement.}
        \label{fig:rwdc_details}
    \end{minipage}
\end{figure*}

Initial works on camera-radar fusion typically follow Region-of-Interest-pooled late-fusion approaches in the perspective or BEV-view \cite{kim2020grif, nabati2021centerfusion, kim2022craft}. 
Others directly concatenate view-aligned radar feature maps to high-level image features and feed the fused features into a standard 3D detector head~\cite{wu2023mvfusion, long2023radiant}.
X3KD \cite{klingner2023x3kd}, BEVGuide \cite{man2023bev} and FUTR3D \cite{chen2022futr3d} employ BEV query-based feature sampler with a transformer decoder.
In general, fusing the radar in its natural 3D representation of the BEV grid has proven to be most effective sofar \cite{yao2023radarsurveycomprehensive, harley2022simple, zhou2023bridging, kim2023crn}.
RCBEV~\cite{zhou2023bridging} empirically shows that using specialized or heavy point-cloud-processing feature backbones does not result in performance gains - hence why our \sota~architecture leverages efficient pillar voxelization \cite{lang2019pointpillars, li2023pillarnext}.

CRN \cite{kim2023crn} introduces a multi-modal deformable attention mechanism, a BEV-fusion feature operator with a larger receptive field than simple concatenation~\cite{liu2022bevfusion}.
While they rely on a monocular depth network and flatten image features to use the radar occupancy as a second view-transformation, we aim to strengthen the depth estimation itself by early-fusion of the two independent feature spaces.
Our approach addresses spatial misalignment from monocular depth cues by moving fusion to a lower level, employing a hybrid fusion.

\section{\hydra\ Architecture}
\label{sec:main}
We introduce \textbf{\name}, a novel camera radar fusion architecture aiming to reduce the depth prediction error and leading to state-of-the-art performance in 3D object detection and semantic occupancy prediction.
The overall architecture is visualized in Figure \ref{fig:teaser} and consists of the following key components:
\begin{itemize}
    \item \textbf{Modality-specific Feature Encoder:} 
    Multi-view images are fed into a 2D encoder and converted to high-level feature maps. The radar point cloud is voxelized and encoded to one height level by point pillars \cite{lang2019pointpillars}.
    \item \textbf{Unified Depth Prediction:} 
    Our newly proposed \hatl\ leverages cross-attention to associate pillar features (missing height, sparse depth) in the respective camera frustrum with image columns (missing depth, dense height). 
    The geometry-aware features are added residually and converted into a dense depth distribution.
    \item \textbf{BEV Fusion:} 
    The forward projection module \cite{huang2021bevdet} generates the initial BEV representation and is concatenated with radar pillar channels and fused by a simple Squeeze-and-Exictation layer \cite{hu2018squeeze, liang2022bevfusion}. 
    \item \textbf{Radar-guided Backward Projection:} 
    The backward projection module~\cite{li2022bevformer} refines the initial sparse features, guided by the \rdc\ calculated from radar BEV features, its implicitly encoded location and radar-aware depth distribution.
    \item \textbf{Downstream Task Heads:} 
    The fused BEV representation is encoded by additional residual blocks and fed into the respective task head.
\end{itemize}

\begin{table*}[!t]
    \centering
    \caption{
        3D Object Detection on nuScenes val set. 
        `L', `C', and `R' represent LiDAR, Camera, and Radar, respectively.
    }
    \label{table:3D det val set}
    \resizebox{0.75\textwidth}{!}{
    \begin{tabular}{l|c|c|c|cc|c@{\hspace{1.0\tabcolsep}}c@{\hspace{1.0\tabcolsep}}c@{\hspace{1.0\tabcolsep}}c@{\hspace{1.0\tabcolsep}}c} 
        \toprule
        \textbf{Methods}                                    & \textbf{Input}    & \textbf{Backbone}     & \textbf{Image Size}   & \textbf{NDS$\uparrow$}    & \textbf{mAP$\uparrow$}    & \textbf{mATE$\downarrow$}     & \textbf{mASE$\downarrow$} & \textbf{mAOE$\downarrow$} & \textbf{mAVE$\downarrow$} & \textbf{mAAE$\downarrow$} \\
        
        \midrule
        CenterPoint-P \cite{yin2021center}                  & L                 & Pillars               & -                     & 59.8                      & 49.4                      & 0.320                         & 0.262                     & 0.377                     & 0.334                     & 0.198 \\
        CenterPoint-V \cite{yin2021center}                  & L                 & Voxel                 & -                     & 65.3                      & 56.9                      & 0.285                         & 0.253                     & 0.323                     & 0.272                     & 0.186 \\
        \midrule
        BEVDepth \cite{li2022bevdepth}                      & C                 & R50                   & $256\times704$        & 47.5                      & 35.1                      & 0.639                         & 0.267                     & 0.479                     & 0.428                     & 0.198 \\
        FB-BEV \cite{li2023fb}                              & C                 & R50                   & $256\times704$        & 49.8                      & 37.8                      & 0.620                         & 0.273                     & 0.444                     & 0.374                     & 0.200 \\
        X3KD \cite{klingner2023x3kd}                        & C+R               & R50                   & $256\times704$        & 53.8                      & 42.3                      & -                             & -                         & -                         & -                         & - \\
        StreamPETR \cite{Wang2023streampetr}                & C                 & R50                   & $256\times704$        & 54.0                      & 43.2                      & 0.581                         & 0.272                     & 0.413                     & 0.295                     & 0.195 \\
        SparseBEV \cite{liu2023sparsebev}                   & C                 & R50                   & $256\times704$        & 54.5                      & 43.2                      & 0.606                         & 0.274                     & \textbf{0.387}            & 0.251                     & 0.186 \\
        CRN \cite{kim2023crn}                               & C+R               & R50                   & $256\times704$        & 56.0                      & 49.0                      & 0.487                         & 0.277                     & 0.542                     & 0.344                     & 0.197 \\
        \gr \hcolor{\hydra}                                 & C+R               & R50                   & $256\times704$        & \textbf{58.5}             & \textbf{49.4}             & \textbf{0.463}                & \textbf{0.268}            & 0.478                     & \textbf{0.227}            & \textbf{0.182} \\
    
        \midrule
        MVFusion \cite{wu2023mvfusion}                      & C+R               & R101                  & $900\times1600$       & 45.5                      & 38.0                      & 0.675                         & 0.258                     & 0.372                     & 0.833                     & 0.196 \\
        FUTR3D \cite{chen2022futr3d}                        & C+R               & R101                  & $900\times1600$       & 50.8                      & 39.9                      & -                             & -                         & -                         & 0.561                     & - \\ %
        CRN \cite{kim2023crn}                               & C+R               & R101                  & $512\times1408$       & 59.2                      & 52.5                      & 0.460                         & 0.273                     & 0.443                     & 0.352                     & \textbf{0.180} \\
        \gr \hcolor{\hydra}                                 & C+R               & R101                  & $512\times1408$       & \textbf{61.7}             & \textbf{53.6}             & \textbf{0.416}                & \textbf{0.264}            & 0.407                     & \textbf{0.231}            & 0.186 \\
        \bottomrule
    \end{tabular}
    }
    
\end{table*}

\subsection{Vision-Centric Foundation}
In our method we propose an extension to existing vision-centric 3D perception systems. Thus we build upon the recent advancements in camera-based high-performance object detection (BEVDet Series \cite{huang2021bevdet,li2022bevdepth}) and the simple but effective concept of fusing radar and image features in the BEV space \cite{harley2022simple}. 
The image encoder with a feature pyramid neck \cite{lin2017feature}, the Context-Net, Depth-Net, and BEV-Encoder \cite{he2016deep} follow the same architecture choices as previous work \cite{huang2021bevdet,li2022bevdepth}.

\subsection{Height Association Transformer}
\label{sec:hat}
Our method utilizes the strength of each modality to overcome the main challenge in vision-centric depth prediction. 
We propose a new transformer-based plug-in module to leverage the complementary radar pillar features (\cf \cref{fig:hat_details}).
The primary challenge is associating the radar features $R$ with the respective image features $F$ in the perspective view. 
We aim to learn which part of the downsampled feature map should benefit from the ambiguous radar encoding with an extended receptive field along the complete image height (in principle reversing the view transformation).

Let $F\in \mathbb{R}^{B \times N \times H \times W \times C}$ denote 2D image backbone output, while $R\in \mathbb{R}^{B \times N \times 1 \times W \times C}$ describes the radar pillar features in the respective camera frustum.
$B$ is the batch size, $N$ the number of cameras, $H$ and $W $the feature height and width, $D$ the number of depth bins in the camera frustum and $C$ the same-size embedding dimension.
For efficiency, we query a single column $w$ of discrete monocular height features $F_{bnw}\in \mathbb{R}^{H \times 1 \times C}$ for the $w$-th respective radar feature keys and values $R_{bnw}\in \mathbb{R}^{1 \times D \times C}$ on the ground plane.
Therefore, we reshape the feature tensors to a new batch size $B'=B\times N\times W$.
For each 'sequence' we apply a learnable positional embedding. 
Next, we encode the height bins with self-attention and decode and fuse the radar features via cross-attention \cite{vaswani2017attention}. 
A peak in activations on the depth bin plane is compared with promising height proposals, encoding the geometric relationship in the attention, inducing sparse but strong metric cues into the dense feature space.
Either modality misses crucial information that the other sensor can provide. 
The resulting feature map $F'$ is the input for the depth and context network, producing the final more robust depth distribution.

\setlength{\tabcolsep}{2mm}
\begin{table}[!t]
\centering
\caption{
    3D Object Detection on the nuScenes test set.
    The ConvNext backbone typically leads to better performance~\cite{li2022bevdepth}.
}\label{table:test}
\resizebox{\linewidth}{!}{
\begin{tabular}{l|c|c|cc|c@{\hspace{1.0\tabcolsep}}c@{\hspace{1.0\tabcolsep}}c@{\hspace{1.0\tabcolsep}}c@{\hspace{1.0\tabcolsep}}c} 
    \toprule
    \textbf{Methods} &                          \textbf{Input}      & \textbf{Backbone}         & \textbf{NDS$\uparrow$}    & \textbf{mAP$\uparrow$}    & \textbf{mATE$\downarrow$} \\
    \midrule
    KPConvPillars \cite{ulrich2022impr}         & R                 & Pillars                   & 13.9                      &  4.9                      & 0.823 \\
    \midrule
    CenterFusion \cite{nabati2021centerfusion}  &C+R                & DLA34                     & 44.9                      & 32.6                      & 0.631 \\
    MVFusion \cite{wu2023mvfusion}              &C+R                & V2-99                     & 51.7                      & 45.3                      & 0.569 \\
    CRAFT \cite{kim2022craft}                   &C+R                & DLA34                     & 52.3                      & 41.1                      & 0.467  \\
    BEVDepth \cite{li2022bevdepth}              & C                 & ConvNeXt-B                & 60.9                      & 52.0                      & 0.445 \\
    CRN \cite{kim2023crn}                       &C+R                & ConvNeXt-B                & 62.4                      & \textbf{57.5}                      & 0.416 \\
    StreamPETR \cite{Wang2023streampetr}        & C                 & V2-99                     & 63.6                      & 55.0                      & 0.479 \\
    SparseBEV \cite{liu2023sparsebev}           & C                 & V2-99                     & 63.6                      & 55.6                      & 0.485 \\           
    \gr \hcolor{\hydra}                         &C+R                & V2-99                     & \textbf{64.2}             & 57.4             & \textbf{0.398} \\
    \bottomrule
\end{tabular}
}
\end{table}

\subsection{Radar-Weighted Depth Consistency}
\label{sec:rwdc}
In our method we tackle two challenges of spatial misalignments and projection inconsistencies when integrating the depth information from multiple sensors. 
Previous approaches \cite{kim2023crn} compensate for spatial misalignment with a global receptive field within the BEV. %
Instead, we propose concatenating the initial set of lifted and splatted BEV features with the radar pillar channels and refining them with our radar-weighted back-projection module.
Extending the back-projection-as-refinement \cite{li2023fb} to the multi-modal domain, we enforce the consistency between the two projection spaces on both sides of the view disparity. %
We upgrade the full cycle of Depth Consistency three-fold to enhance the projection quality, leveraging the synergy of our \hats\ module and a lightweight radar guidance network (RGN). 
With a $3 \times 3$ convolution followed by a sigmoid, the RGN encodes the radar-only BEV features to additional attention weights~$r$ (\cf \cref{fig:rwdc_details}). 

The overall concept relies on the classic projection of 3D BEV points $(x,y,z)$ onto 2D image coordinates $(u,v)$ with projection matrix $P$. 
Every BEV proposal location $Q_{x,y}$ implicitly encodes a depth value $d_Q$ for the corresponding camera image projection. 
\begin{equation}
\label{3to2_projection}
d\cdot\begin{bmatrix} u &v & 1 \end{bmatrix}^T= P \cdot \begin{bmatrix} x &y &z & 1 \end{bmatrix}^T
\end{equation}
By weighting the cross-attention \cite{zhu2021deformable} with the consistency $w_C$ (scalar product) of the predicted depth distribution $d_P$ and implicit query depth value $d_Q$ (converted to a distribution), we enrich the otherwise sparse, misaligned, and especially unmatched single-modal features.
\begin{equation}
    w_c = d_P \cdot d_Q
    \end{equation}
Our \hats\ module tackles the left side of Equation \ref{3to2_projection} by strengthening the explicit depth (pushing effect \cite{harley2022simple} of radar). 
BEV-fusion activates better proposal locations $Q_{x,y}$ for the right side of Equation \ref{3to2_projection} (pulling effect \cite{harley2022simple} of radar).
The deformable cross-attention brings them closer together and enforces depth consistency on both sides of the equation.
We follow the notation of \cite{li2023fb} and adapt to the radar-weighted Spatial Cross-Attention~($\text{SCA}$):~  
\begin{equation}\label{sca_r}
\resizebox{\linewidth}{!}{%
    $\text{SCA}(Q_{x,y}, F)\!= \!\sum\limits_{i=1}^{{N_\text{c}}} \sum\limits_{j=1}^{{N_\text{ref}}}
\mathcal{F}_{d}(Q_{x,y}, \mathcal{P}_i(x, y, z_j), F_i)\cdot\boldsymbol{w_c^{ij}}\cdot\boldsymbol{r_{x,y}},$
}
\end{equation}

with $\mathcal{F}_{d}$ denoting the deformable attention, dynamically sampling features around the projection point $\mathcal{P}_i(x, y, z_j)$ on the image feature map $F_i$, weighted by depth consistency $w_c$ and radar attention $r$.
Equation \ref{sca_r} highlights where the complementary hybrid fusion approach is beneficial for a more robust depth prediction. 
Furthermore, we moved the backprojection after the temporal fusion to capitalize on partially occluded objects indicated not only by radar reflections but also propagated history features.
\subsection{Down-stream Tasks}
\label{sec:down}
Our final BEV representation can be leveraged for multiple 3D perception tasks, generalizing well for 3D object detection, 3D multi-object tracking (MOT), and 3D semantic occupancy prediction.
To decode the features into 3D bounding boxes characteristics, we follow standard practices and use a simple anchor-free center-based head \cite{yin2021center}.
Similar to \cite{kim2023crn}, we utilize the same box-based tracking-by-detection approach of CenterPoint \cite{yin2021center}, which relies on a velocity-based greedy distance matching. 
Our powerful fused BEV features can be directly converted into a rich semantic occupancy output without the need to forward projecting features into the full 3D voxel cube or a 3D-convolution-based BEV encoder.
We use a one-layer 1$\times$1 Channel-To-Height convolution \cite{yu2023flashocc} to create the final representation for our occupancy head \cite{li2023fbocc}. 
We simply enlarge the final BEV features $F\in \mathbb{R}^{B \times X \times Y \times C}$ in the channel dimension $C$ to $C' = Z \times C$ and reshape the feature map to the new height dimension $Z$, resulting in $F'\in \mathbb{R}^{B \times X \times Y \times Z \times C}$. 
This new unstacked representation is fed into the low-cost occupancy head \cite{li2023fbocc}.

\section{Experiments}
\label{sec:experiments}
\subsection{Dataset and Metrics}
We perform extensive experiments and ablations on nuScenes \cite{caesar2020nuscenes}, View-of-Delft \cite{palffy2022multi} and Occ3D~\cite{tian2024occ3d}. %
The former two provide a rich set of metrics to assess the quality of 3D perception algorithms:
mean Average Precision \cite{everingham2010pascal}, mean Average Translation, Scale, Orientation, Velocity, and Attribute Error. The NDS is a weighted average of all other metrics.
The AMOTA score~\cite{weng2019baseline} evaluates the overall 3D multi object tracking performance.
On Occ3D the goal is to predict the complete 3D scene geometry, consisting of a voxelized representation for 18 classes. 
The benchmark reports the mean IoU over the number of true positive, false positive, and false negative voxel predictions.

\subsection{Implementation Details} 
Following the literature, we train our model in three different scaling settings.
We set the backbones to ResNet50~\cite{he2016deep} with an input resolution of 256$\times$704, 118 depth categories, and a pillar and BEV grid size of 0.8m, leading to 128$\times$128 BEV space.
Scaling up the model backbones and image resolutions to ResNet101~\cite{he2016deep} (512$\times$1408) and V2-99~\cite{lee2019vov} (640$\times$1600), the BEV space is doubled to a standard of 256$\times$256. 
For comparison to the latest camera-based methods on the test set, we opt for the smaller and more efficient V2-99 backbone.

We build upon the code base of the BEVDet Series~\cite{huang2021bevdet} and train for 20 epochs following the memory-efficient sequential sampling  of \cite{park2022time} without CBGS \cite{zhu2019class}. 
This reduces the step-to-step sampling diversity but speeds up the training time significantly. 
To trade off the number of parameters, we half the backbone output and depth net middle channels.
Our \hats\ module is implemented by one layer of self- and cross-attention working on 16-times downsampled height queries. This leads to sequences of 16, 32, and 40 height bins for the three model sizes. For temporal fusion, we follow \cite{li2023fb,liu2023sparsebev, Wang2023streampetr} and use the last eight history frames.
Data augmentations are the same as BEVDet \cite{huang2021bevdet} and CRN~\cite{kim2023crn}, but no test-time augmentations are used.
For a fair evaluation, each model is trained only on the respective task and leverages the same task head and supervision signals as the baseline methods.
\subsection{Main Results}
\begin{figure*}[t]
	\centering
	\includegraphics[width=0.8\textwidth]{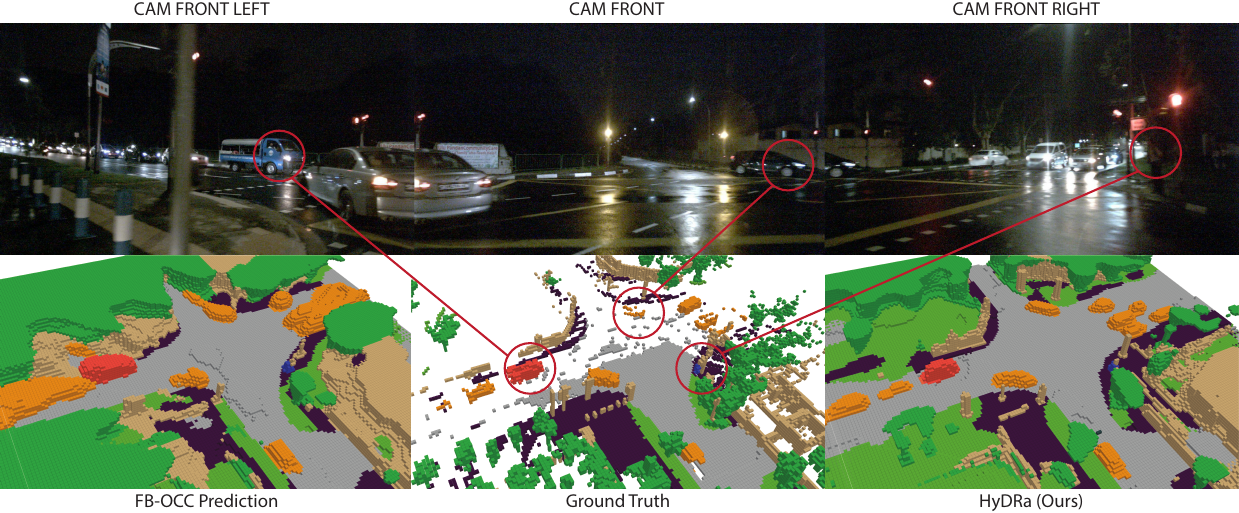}

	\caption{Qualitative comparison of the semantic occupancy prediction in a challenging night scenario. 
	The top row shows the front-view input cameras. 
	We compare FB-OCC \cite{li2023fbocc} with our proposed \name. 
	While the baseline struggles to distinguish different objects at distance, HyDRa showcases spatial consistency and robustness of the detected cars (orange), truck (red) and pedestrian (blue).
	}
	\label{fig:occ}
\end{figure*} %

\textbf{3D Object Detection.}
In Table \ref{table:3D det val set} we report the results of our \name\ model on the nuScenes validation set. With a ResNet50 backbone, we achieve a new \sota\ of 58.5 NDS, surpassing the previous best score of 56.0 NDS by CRN \cite{kim2023crn} by 2.5 NDS. 
Scaling the model and image resolution to ResNet101 and 512$\times$1408, we maintain the large margin and increase the NDS to 61.7, outperforming all other camera- and radar-based methods on the val split. We want to highlight the impressive improvement in the mean Average Translation Error (mATE) of 0.416, which can be attributed to our model's robust depth estimation concept.
Our \name\ surpasses CRN with the more powerful ConvNext-Base\cite{li2022bevdepth} backbone by 1.4 NDS due to strengthened translation and velocity estimation, leading to a new \sota\ of 64.2 NDS on the nuScenes test set (see Table \ref{table:test}). 

Additionally, we show in Table \ref{table:vod} that HyDRa generalizes on the recent View-of-Delft validation set \cite{palffy2022multi}, utilizing a more advanced 4D millimeter-wave radar. 
HyDRa outperforms the best method by an increase of +4.57 mAP in the entire annotation area and +4.06 mAP in the region of interest (compared to the CRN-inspired LXL model \cite{xiong2023lxl}).\\

\noindent \textbf{3D Multi Object Tracking.}
We summarize the tracking results in Table \ref{table:3D Track test set}. Benefitting from accurate object localization and high-quality velocity estimation, the center-based tracking performance of \name\ is competitive with the state-of-the-art methods, leading to the best trade-off between false positives (FP), false negatives (FN), and identity switches (IDS) on the nuScenes test set, once more narrowing the gap to LiDAR-based CenterPoint. \\

\setlength{\tabcolsep}{2mm}
\begin{table}[t!]
    \centering
    \caption{
        3D object detection results on View-of-Delft ~\cite{palffy2022multi}. The official eval settings differentiates a Region of interest.
    }
    \resizebox{\linewidth}{!}{
    \begin{tabular}{l|c|ccc|c|ccc|c}
        \midrule
        \multirow{2}{*}[-0.7ex]{\textbf{Methods}} & \multirow{2}{*}[-0.7ex]{\textbf{Input}}  & \multicolumn{4}{c|}{\textbf{AP Entire Annotated Area (\%)}} & \multicolumn{4}{c}{\textbf{AP Region of Interest (\%)}} \\
        \vspace{-0.8em} \\ %
        \cline{3-10}
        \vspace{-0.8em} \\ %
        & & Car & Ped. &Cyclist& mAP & Car & Ped.& Cyclist& mAP \\
        \midrule
        PointPillar~\cite{lang2019pointpillars} & R &37.1& 35.0& 63.4& 45.2& 70.2& 47.2& 85.1& 67.5 \\
        RCFusion \cite{zheng2023rcfusion} & C+R &41.7& 39.0& 68.3& 49.7& 71.9& 47.5& 88.3& 69.2 \\ %
        SMURF~\cite{yao2023smurf} & R& 42.3 &39.1 &71.5 &51.0 &71.7& 50.5& 86.9& 69.7\\
        LXL \cite{xiong2023lxl} & C+R &42.3& 49.5 &77.1& 56.3& 72.2& 58.3 &88.3 &72.9 \\ %
        \gr \hcolor{HyDRa} & C+R & 52.8 & 56.6 & 73.3 & \textbf{60.9} & 80.7 & 62.9 & 87.4 & \textbf{77.0}  \\
        \bottomrule
    \end{tabular}
    }

    \label{table:vod}
\end{table}

\renewcommand{\thefootnote}{2}
\setlength{\tabcolsep}{2mm}
\begin{table}[!t]
\centering
\caption{
    3D Object Tracking on nuScenes test set.
}\label{table:3D Track test set}
\resizebox{\columnwidth}{!}{
\begin{tabular}{l|c|c|cc|ccc}
    \toprule
    \textbf{Methods}                                        & \textbf{Input}    & \textbf{Backbone}     & \textbf{AMOTA$\uparrow$}  & \textbf{AMOTP$\downarrow$}    & \textbf{FP$\downarrow$}   & \textbf{FN$\downarrow$}   & \textbf{IDS$\downarrow$}\\
    \midrule
    CenterPoint \cite{yin2021center}                        & L                 & Voxel                 & 63.8                      & 0.555                         & 18612                     & 22928                     & 760 \\
    \midrule
    UVTR \cite{li2022unifying}                              & C                 & \small{V2-99}         & 51.9                      & 1.125                         & 14994                     & 39209                     & 2204 \\
    ByteTrackV2 \cite{zhang2023bytetrackv2}                 & C                 & \small{V2-99}         & 56.4                      & 1.005                         & 18939                     & 33531                     & \textbf{704} \\
    StreamPETR \cite{yang2022quality}   & C                 & \small{ConvNeXt-B}    & 56.6                      & 0.975                         & 21268                     & \textbf{31484}            & 784 \\ %
    CRN  \cite{kim2023crn}                                  & C+R               & \small{ConvNeXt-B}    & 56.9                      & \textbf{0.809}                & 16822                     & 41093                     & 946 \\ %
    \gr \hcolor{\hydra}                                     & C+R               & \small{V2-99}         & \textbf{58.4}             & 0.950                         & \textbf{13996}            & 32950                     & 791 \\ %
    \bottomrule
\end{tabular}
}
\end{table}

\begin{table}[!t]
\centering
\caption{
3D Occupancy Prediction on Occ3D (CVPR2023 Occupancy Challenge \cite{tian2024occ3d, tong2023scene}). 
The mIoU based on 16 different classes.
All models use a r50 backbone on a resolution of $256\times704$.
}\label{table:occ_result}
\setlength{\tabcolsep}{2mm}
\resizebox{\linewidth}{!}{
\begin{tabular}{l|c|c|ccccc}
    \toprule
    \textbf{Methods}                        & \textbf{Input}        & \textbf{mIoU $\uparrow$}  & \textbf{Bus}  & \textbf{Car}  & \textbf{Ped.} & \textbf{Truck}  & \textbf{MC}\\ %
    \midrule
    SparseOcc \cite{liu2023fully}           & C                      & 30.9                      & 32.9          & 43.3          & 23.4 & 29.3 & 23.8 \\
    BEVStereo-OCC \cite{huang2021bevdet}    & C                      & 36.1                      & 42.1          & 49.6          & 21.5 & 37.1 & 17.4 \\
    FlashOcc (M5) \cite{yu2023flashocc}     & C                      & 39.0                      & 41.6          & 50.5          & 23.8 & 37.4 & 23.0\\ %
    RenderOcc \cite{pan2023renderocc}       & C                      & 39.7                      & -             & -             & -  & - & - \\ %
    FB-OCC \cite{li2023fb, li2023fbocc}     & C                      & 40.7                      & 45.3          & 50.5          & 28.2 & 37.9 & 29.0\\
    \gr \hcolor{\hydra}                     &C+R                     & \textbf{44.4}             & \textbf{52.3}  & \textbf{56.3} & \textbf{35.1} & \textbf{44.1} & \textbf{35.9} \\
    \bottomrule
\end{tabular}
}
\end{table}

\noindent \textbf{3D Occupancy Prediction.}
Dense architectures with strong depth prediction excel in occupancy prediction \cite{tian2024occ3d,tong2023scene}. 
Strengthening the most relevant bottlenecks, HyDRa showcases superior performance on Occ3D, outperforming all other camera-based methods by a large margin of 3.7 mIoU (see Table \ref{table:occ_result}). 
It works remarkably well on challenging dynamic objects as visualized in \cref{fig:occ}. 
Complementary camera-radar fusion is the key to unlocking the full potential of vision-centric 3D sensing.  
\subsection{Ablation Studies}

\begin{table}[!t]
    \centering
    \caption{
    Ablation of fusion strategies:
    Baseline Range-Fusion (RF); 
    CRN with occupancy-based Radar View Transformation (RVT); 
    We train and evaluate in the front-view setting.
    }
    \label{table:combined_ablations}
    \resizebox{\linewidth}{!}{
    \setlength{\tabcolsep}{2mm}
    \begin{tabular}{l|c|c c c|c c c c}
        \toprule
        \textbf{Methods}                            & \textbf{Input}      &\textbf{BEVF}    &\textbf{4D}      &\textbf{Depth}        &\textbf{NDS$\uparrow$}    &\textbf{mAP$\uparrow$}       & \textbf{mATE$\downarrow$}   & \textbf{mAVE$\downarrow$} \\
        \midrule
        BEVDepth \cite{li2022bevdepth}              & C                   & \xmark          & \cmark          & \cmark               & 44.4                    & 32.1                       & 0.737                       & 0.344    \\
        \midrule
        BEVDepth + RVT \cite{kim2023crn}                   & C+R                 & \xmark          & \cmark          & \cmark               & 45.6                    & 38.3                       & 0.619                       & 0.510    \\ 
        BEVDepth + RF                               & C+R                 & \xmark          & \cmark          & \cmark               & 48.1                    & 37.2                       & 0.614                       & \textbf{0.284}     \\
    \gr BEVDepth + HAT                              & C+R                 & \xmark          & \cmark          & \cmark               & \textbf{50.3}           & \textbf{39.8}              & \textbf{0.586}              & 0.287 \\
        \midrule
        CRN incl. RVT \cite{kim2023crn}                       & C+R                 & \cmark          & \cmark          & \cmark               & 48.9                    & 39.7                       & 0.616                       & 0.361    \\
        BEVDepth + RDC                                        & C+R                 & \cmark          & \cmark          & \cmark               & 52.7                    & 40.5                       & 0.534                       & 0.261 \\
        \gr \hcolor{\hydra} (RDC + HAT)                      & C+R                 & \cmark          & \cmark          & \cmark               & \textbf{53.6}           & \textbf{41.2}              & \textbf{0.493}              & \textbf{0.257} \\
        \bottomrule
    \end{tabular}
    }
    \end{table}

To validate the contribution of our proposed designs, we conducted detailed ablation studies on nuScenes (val set).
The training of \cref{table:combined_ablations} and \cref{table:singlesetting} is performed in the front-view setting. %
Input for the model are the three forward-facing cameras and the corresponding radar sensors, reducing the data and computation size by a factor of two.\\

\noindent \textbf{Effect of the Range View Fusion.} 
We develop a simple baseline similar to Radiant \cite{long2023radiant} to show that using radar in a unified depth net brings more performance than separate branches. %
We extend BEVDepth \cite{li2022bevdepth} with an additional ResNet18 \cite{he2016deep} encoding the radar in the same image view. By projecting radar points into the image plane, we create a range image of the radar point cloud, converting 3D points into a pixel location. 
Instead of RGB channels, radar channels are used, with every non-radar pixel set to zero. 
This pseudo-image is input to a smaller ResNet18 encoder, and each output stage is concatenated to the respective RGB encoder output. 

We show in Table \ref{table:combined_ablations} that this unified approach already outperforms the RVT of CRN. We indicate, that the velocity estimation is also benefitting from a perspective fusion, compared to pure radar-occupancy-guided transformation of the semantic features. 
However, processing the mostly sparse stripe of radar points with 2D convolutions is neither effective nor efficient.
To overcome the limitations of the naive baseline, we propose a more sophisticated approach, the \hats\ module, which is a novel way of fusing radar and camera features in the perspective view and is capable of significantly improving the depth estimation of camera-based models.
Compared to related work, we do not have to rely on heuristic height extensions \cite{long2023radiant,nabati2021centerfusion} or collapsing image features \cite{kim2023crn}. \\

\noindent \textbf{Effect of Radar Weighted Depth Consistency.} 
Radar-informed depth consistency is a crucial concept of our architecture, showing significant and consistent improvements over the baseline architecture of CRN. 
Especially adding the \hats\ module into this paradigm leverages synergistic effects, leading to strong metric depth understanding (\cf \cref{table:combined_ablations}). 
Without temporal or depth supervision, as ablated in Table \ref{table:singlesetting}, \name\ is the superior concept for increased depth and velocity sensing, showing a decent improvement over the camera-only baseline. 
This could be important for time-critical situations e.g. collision avoidance \cite{yao2023radarsurveycomprehensive}.

\begin{table}[!t]
    \centering
    \caption{
    Ablation without temporal fusion or depth supervision. We train and evaluate in the front-view setting.
    }
    \label{table:singlesetting}
    \setlength{\tabcolsep}{2mm}
    \resizebox{\linewidth}{!}{
    \begin{tabular}{l|c|c c|c c c c}
        \toprule
        \textbf{Methods}                            & \textbf{Input}        & \textbf{4D}       & \textbf{Depth}         & \textbf{NDS$\uparrow$}       &\textbf{mAP$\uparrow$}         & \textbf{mATE$\downarrow$}     & \textbf{mAVE$\downarrow$} \\
        \midrule
        BEVDepth \cite{li2022bevdepth}              & C                     & \xmark            & \xmark                 & 33.3                        & 26.7                         & 0.803                         & 0.930 \\
        CRN \cite{kim2023crn}                       & C+R                   & \xmark            & \xmark                 & 42.5                        & \textbf{35.0}                & 0.607                         & 0.584 \\ %
        \gr \hcolor{\hydra}                         & C+R                   & \xmark            & \xmark                 & \textbf{45.8}               & 34.7                         & \textbf{0.569}                & \textbf{0.482} \\
        \bottomrule
    \end{tabular}
    }
    
    \end{table}

\section{Conclusion}
\label{sec:conclusion}
We introduce \textbf{\hydra}, the new \sota\ hybrid fusion paradigm, excelling in various 3D perception tasks and showing a promising path for future research in the field of radar-based detection.
\name\ beats the former \sota\ in camera-radar fusion by a clear margin of 1.8 NDS and naturally generalizes to the first radar-enhanced occupancy prediction model, improving upon the best camera baseline by 3.7 mIoU.
With this line of work, we want to contribute to safer autonomous driving due to better handling of low-visible objects, more robust depth estimation, and accurate velocity estimation.

\bibliographystyle{IEEEtran}
\bibliography{IEEEabrv,root}

\newpage
\appendix
\section{Appendix}

\label{sec:overview}
In this supplementary material, we highlight additional information about our proposed \hydra~, a novel 3D scene understanding framework that leverages the complementary strengths of camera and radar sensors.
In \cref{sec:additional} and \cref{sec:qualitative}, we present further qualitative and quantitative results in occupancy prediction, respectively, to provide a holistic understanding of the significant improvements in performance unlocked by \hydra~'s architecture.

\subsection{Additional Evaluation Results}
\label{sec:additional}

\noindent \textbf{Adverse Conditions.}
In Table \ref{table:Occ_val_set}, we investigate the performance of \hydra~ under challenging environmental conditions.
We focus on the task of occupancy prediction to generalize the findings of \cite{kim2023crn} to complete 3D scene understanding.
We report the mIoU of different subsets of the validation set, focusing on scenarios with night and rain conditions.
We observe that \hydra~ outperforms the baseline method FB-OCC \cite{li2023fb, li2023fbocc} (public checkpoint (E)) in all conditions. 
The improvement is particularly significant in the night and night \& rain conditions, where \hydra~ achieves a 16.7\% and 33.5\% improvement, respectively.
The consistency in improvement across different conditions demonstrates the perception robustness of radar in adverse conditions, in general, and \hydra~ in particular.\\

\begin{table}[h]
\centering
\caption{
3D Occupancy Prediction on different lighting and weather conditions of Occ3D \cite{tian2024occ3d} ((nuScenes val set), mIoU $\uparrow$). 
}
\resizebox{\columnwidth}{!}{
\begin{tabular}{l|c|c|cccc|c}
    \toprule
    \textbf{Methods}                        & \textbf{Input}       & \textbf{All}              & \textbf{Sunny}    & \textbf{Rainy}    & \textbf{Day}      & \textbf{Night}        & \textbf{Night \& Rain}\\ %
    \midrule
    FB-OCC \cite{li2023fb, li2023fbocc}     & C              & 39.1                      & 39.1             &  40.6            & 40.16             & 25.8                  &  22.7 \\
    \gr \hcolor{\hydra}                     &C+R          & \textbf{44.4}            & \textbf{44.0}         & \textbf{45.6}             & \textbf{45.1}             & \textbf{30.1}                 &  \textbf{30.3}  \\
    Improvement                              & -           & +13.5\%                   & +12.5\%          & +12.3\%          & +12.2\%           & +16.7\%               & +33.5\% \\
    \bottomrule
\end{tabular}
}
\label{table:Occ_val_set}
\end{table}

\begin{table*}[!h]
    \centering
    \caption{\parbox{\textwidth}{\centering
        3D Occupancy Prediction on Occ3D \cite{tian2024occ3d} based on nuScenes \texttt{val} set, including IoU of all 16 classes contributing to the mIoU.
    }}
    \setlength{\tabcolsep}{1mm}
    \begin{tabularx}{0.98\textwidth}{l|*{17}{X}|c}
    \toprule
    Method                                      & \rotatebox{90}{others} & \rotatebox{90}{barrier}   & \rotatebox{90}{bicycle} & \rotatebox{90}{bus}     & \rotatebox{90}{car}    & \rotatebox{90}{const. veh.} & \rotatebox{90}{motorcycle} & \rotatebox{90}{pedestrian} & \rotatebox{90}{traffic cone} & \rotatebox{90}{trailer} & \rotatebox{90}{truck}  & \rotatebox{90}{driv. surface} & \rotatebox{90}{other flat} & \rotatebox{90}{sidewalk} & \rotatebox{90}{terrain} & \rotatebox{90}{manmade} & \rotatebox{90}{vegetation} & \rotatebox{90}{mIoU} \\
    \midrule
    TPVFormer \cite{huang2023tri}               & 7.2                   & 38.9                     & 13.7                 & 40.8                     & 45.9                                   & 17.2                 & 20.0                  & 18.9                   & 14.3                  & 26.7                 & 34.2                   & 55.7                   & 35.5                  & 37.6                   & 30.7                  & 19.4                   & 16.8                  & 27.8          \\
    BEVFormerOcc \cite{li2022bevformer}         & 5.9                   & 37.8                     & 17.9                 & 40.4                     & 42.4                                   & 7.4                  & 23.9                  & 21.8                   & 21.0                  & 22.4                 & 30.7                   & 55.4                   & 28.4                  & 36.0                    & 28.1                  & 20.0                   & 17.7                  & 26.9          \\
    CTF-Occ  \cite{tian2024occ3d}               & 8.1                   & 39.3                     & 20.6                 & 38.3                     & 42.2                                   & 16.9                 & 24.5                  & 22.7                   & 21.1                  & 23.0                 & 31.1                   & 53.3                   & 33.8                  & 38.0                   & 33.2                  & 20.8                   & 18.0                   & 28.5          \\
    SparseOcc \cite{liu2023fully}               & 10.6                   & 39.2                      & 20.2                  & 32.9                  & 43.3                                    & 19.4                  & 23.8                   & 23.4                    & 29.3                   & 21.4            & 29.3                    & 67.7                    & 36.3                   & 44.6                    & 40.9                   & 22.0                    & 21.9                   & 30.9           \\
    BEVStereo-OCC \cite{huang2021bevdet}        & 8.2                   & 44.2                     & 10.3                 & 42.1                     & 49.6                                   & 23.4                 & 17.4                  & 21.5                   & 19.7                   & 31.3                 & 37.1                   & 80.1                   & 37.4                  & 50.4                   & 54.3                  & 45.6                   & 39.6                  & 36.0          \\
    OctreeOcc  \cite{lu2023octreeocc}           & -                      & -                        & -                     & -                          & -                                   & -                 & -                        & -                     & -                      & -                      & -                        & -                     & -                        & -                      & -                        & -                      & -                         & 37.4           \\             
    FlashOcc (M5) \cite{yu2023flashocc}         & 11.0                   & 47.2                    & 22.2                 & 41.4                     & 50.6                                   & 23.2                 & 23.0                  & 23.8                   & 26.2                  & 32.1                 & 37.4                   & \textbf{82.2}                   & 42.5                  & 53.3                   & 56.6                  & 46.7                    & 39.7                  & 38.8          \\
    RenderOcc \cite{pan2023renderocc}           & -                      & -                        & -                     & -                          & -                                  & -                  & -                        & -                     & -                      & -                      & -                        & -                     & -                        & -                      & -                        & -                      & -                         & 39.7           \\    
    FB-OCC \cite{li2023fb, li2023fbocc}         & 14.4                  & 45.8                     & 29.2                 & 45.3                     & 50.5                                   & 27.9                 & 29.0                  & 28.2                   & 28.6                  & 32.9                 & 37.9                   & 81.8                   & \textbf{45.5}                  & \textbf{54.0}                   & \textbf{58.7}                 & 43.5                   & 38.8                  & 40.7          \\ 
    \gr \hcolor{\hydra}                         & \textbf{15.1}        & \textbf{51.1}             & \textbf{32.7}         & \textbf{52.3}            & \textbf{56.3}                          & \textbf{29.4}         & \textbf{35.9}          & \textbf{35.1}            & \textbf{33.7}           & \textbf{39.1}         & \textbf{44.1}           & 80.4          & 45.1                           &52.0           & 55.3          & \textbf{52.05}           & \textbf{44.4}          & \textbf{44.4}  \\
    Difference                                  & +0.7                  & +5.3                     & +3.5                 & +7.0                     & +5.8                                   & +1.5                 & +6.9                  & +6.9                   & +4.8                  & +6.2                 & +6.2                   & -1.4                   & -0.4                  & -2.0                   & -3.4                  & +8.7                   & +5.6                  & +3.7          \\
    \bottomrule
\end{tabularx}

\label{tab:occ_detailed}
\end{table*}

\noindent \textbf{Per-Class Analysis.}
To provide a detailed overview of the performance of \hydra~ in 3D scene understanding, we report the IoU of all 16 classes contributing to the mIoU in Table \ref{tab:occ_detailed}.
Occ3D is particularly well suited to evaluate semantic and geometric expressiveness of 3D scene understanding methods, as it provides variety of general objects that are not limited to countable traffic participants.
Details of irregular object shapes and sizes such as trailers, buses, and construction vehicles, are particularly challenging.
Traditionally 3D object detection predicts compact 3D bounding boxes, whereas 3D semantic occupancy prediction estimates the full extent of the object, including semantic class and occupancy status of every voxel within the surrounding environment.
We compare \hydra~ with state-of-the-art methods on the Occ3D \cite{tian2024occ3d} \cite{tong2023scene} and observe that \hydra~ outperforms all methods in terms of mIoU.
This emphasizes the benefit of radar-based perception for holistic scene understanding.
The overall improvement can be attributed to dynamic objects such as cars and motorcycles (metallic objects with high reflectivity for radar sensors), but also pedestrians and cyclists. 
HyDRa also outperforms other methods for large and irregular objects such as buses, trucks, and trailers, which are particularly challenging for 3D object detection methods. 
The already high detection quality of the drivalable area cannot be further improved by radar fusion, whereas vegetation and gneral manmade objects increase in perception quality.

\subsection{Qualitative Results}
\label{sec:qualitative}

\begin{figure*}[h!]
	\centering
	\includegraphics[width=0.98\textwidth]{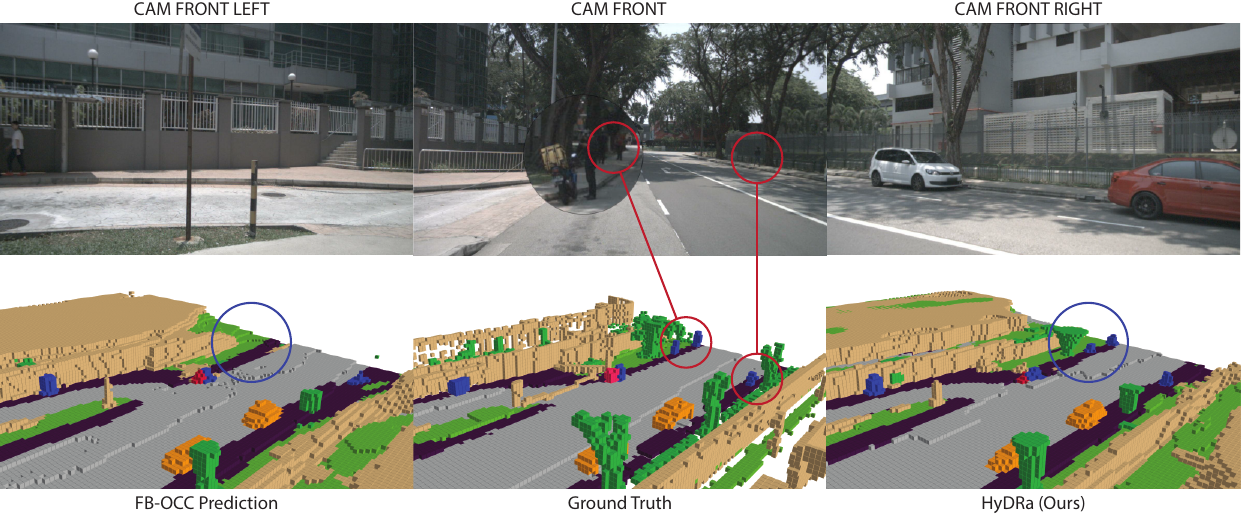}

	\caption{Qualitative comparison of the semantic occupancy prediction under difficult lighting conditions. 
	The top row shows the front-view input cameras. 
	\name is able to to detect two distant pedestrians (blue) next to a tree (green), whereas they are barely distinguishable from the background and not detected by the camera-based baseline.
	}
	\label{fig:occ2}
\end{figure*}

\begin{figure*}[h!]
	\centering
	\includegraphics[width=0.98\textwidth]{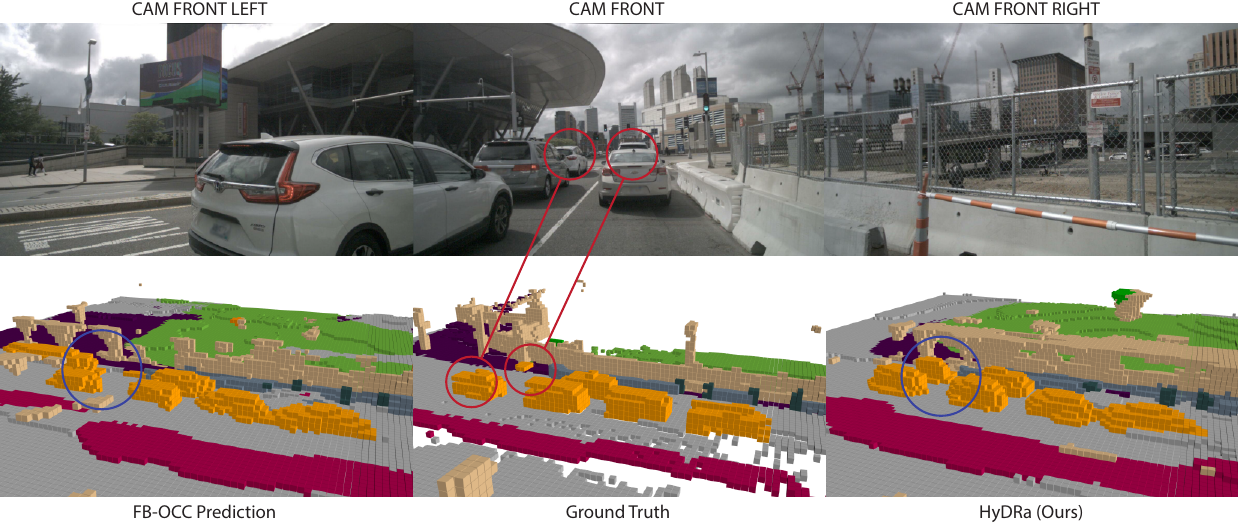}

	\caption{Qualitative comparison of the semantic occupancy prediction in crowded urban scenarios.
	The top row shows the front-view input cameras. 
	While FB-OCC \cite{li2023fbocc} struggles with partially visible and occluded objects, \hydra is able to outperform(lidar-based) ground truth.
	}
	\label{fig:occ3}
\end{figure*}

In this section, we provide additional qualitative results to showcase the robustness and generalization capabilities of \hydra~ in challenging scenarios.
We compare \hydra~ with the baseline method FB-OCC \cite{li2023fbocc} in crowded urban scenarios(\cf \cref{fig:occ3}), challenging lighting conditions (\cf \cref{fig:occ2}), and adverse weather conditions (\cf \cref{fig:occ}).

\end{document}